\documentclass[lettersize,journal]{IEEEtran}
\usepackage{amsmath,amsfonts}
\usepackage{algorithmic}
\usepackage{algorithm}
\usepackage{array}
\usepackage[caption=false,font=normalsize,labelfont=sf,textfont=sf]{subfig}
\usepackage{textcomp}
\usepackage{stfloats}
\usepackage{url}
\usepackage{verbatim}
\usepackage{graphicx}
\usepackage{cite}

\begin{document}

\title{A Survey on Adversarial Robustness of LiDAR-based Machine Learning Perception in Autonomous Vehicles}

\author{ Junae Kim, Amardeep Kaur  
}



\maketitle

\begin{abstract}
In autonomous driving, the combination of AI and vehicular technology offers great potential. However, this amalgamation comes with vulnerabilities to adversarial attacks. This survey focuses on the intersection of Adversarial Machine Learning (AML) and autonomous systems, with a specific focus on LiDAR-based systems. We comprehensively explore the threat landscape, encompassing cyber-attacks on sensors and adversarial perturbations. Additionally, we investigate defensive strategies employed in countering these threats. This paper endeavors to present a concise overview of the challenges and advances in securing autonomous driving systems against adversarial threats, emphasizing the need for robust defenses to ensure safety and security.
\end{abstract}


\section{Introduction}\label{1Introduction}
\IEEEPARstart{I}{n} the field of autonomous driving systems, the combination of Artificial Intelligence (AI) and vehicular technology has opened new possibilities. These advanced systems enable unmanned vehicles, including drones and other autonomous platforms, to not only perceive their surroundings but also make real-time decisions while navigating complex traffic scenarios. With the potential to revolutionize various sectors, autonomous systems promise heightened safety, streamlined logistics, and the capacity for multitude of beneficial operations.

However, beneath the promise of seamless automation lies a significant challenge: the vulnerability of these systems to adversarial attacks, which encompass a wide range of techniques, including those that share characteristics with Adversarial Machine Learning (AML) attacks. AML attacks, designed to manipulate machine learning models, typically involve the introduction of carefully crafted perturbations or alterations to input data. However, it is important to note that perturbing inputs represent just one facet of AML attacks~\cite{Vassilev2024}. The diverse AML landscape includes other techniques like backdoor data poisoning~\cite{Jiang2020}, where an adversary can inject a small number of poisoned samples with a backdoor trigger into the training data. When activated during deployment, these triggers manipulate the machine learning (ML) model. These alterations may be imperceptible to humans but can significantly disrupt the operation of the ML system, potentially causing malfunction. 

The shared characteristics between adversarial threats and AML attacks include the ability to add noise or manipulate minor parts of inputs in ways unrecognizable by humans, with the ultimate target being the ML system, leading to its malfunction. By exploiting vulnerabilities in the system, adversarial attacks aim to create inputs that appear legitimate but are subtly altered to trick the system. This survey paper sets out to explore the intersection of adversarial attacks from an AML perspective and ML-based autonomous driving, addressing a significant subject in this domain.

\begin{figure}[!t]
	\centering
	\includegraphics[width=3.5in]{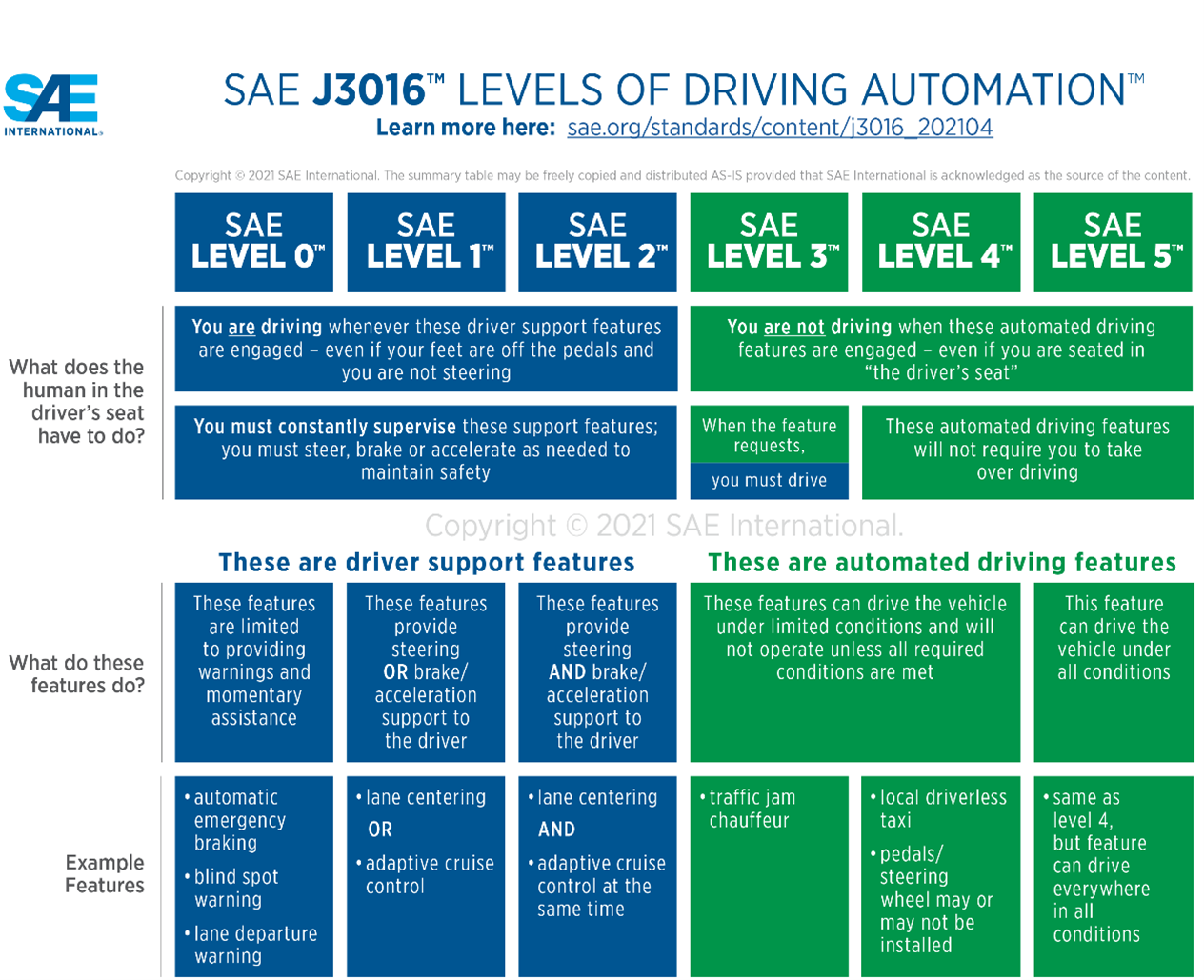}
	\caption{SAE Driving Automation Levels~\cite{SAE2021}}
	\label{fig_1}
\end{figure}

The autonomous driving systems are categorized into six levels by the Society of Automotive Engineers~\cite{SAE2021}, ranging from Level 0 indicating no automation to Level 5 representing complete automation, as illustrated in Figure~\ref{fig_1}. While Level 5 automation is yet to be achieved, ongoing testing efforts are underway~\cite{Khan2023}. Legal constraints in numerous countries also limit the testing and deployment of Autonomous Vehicles (AV).

\begin{figure*}[!t]
	\centering
	\includegraphics[width=7.0in]{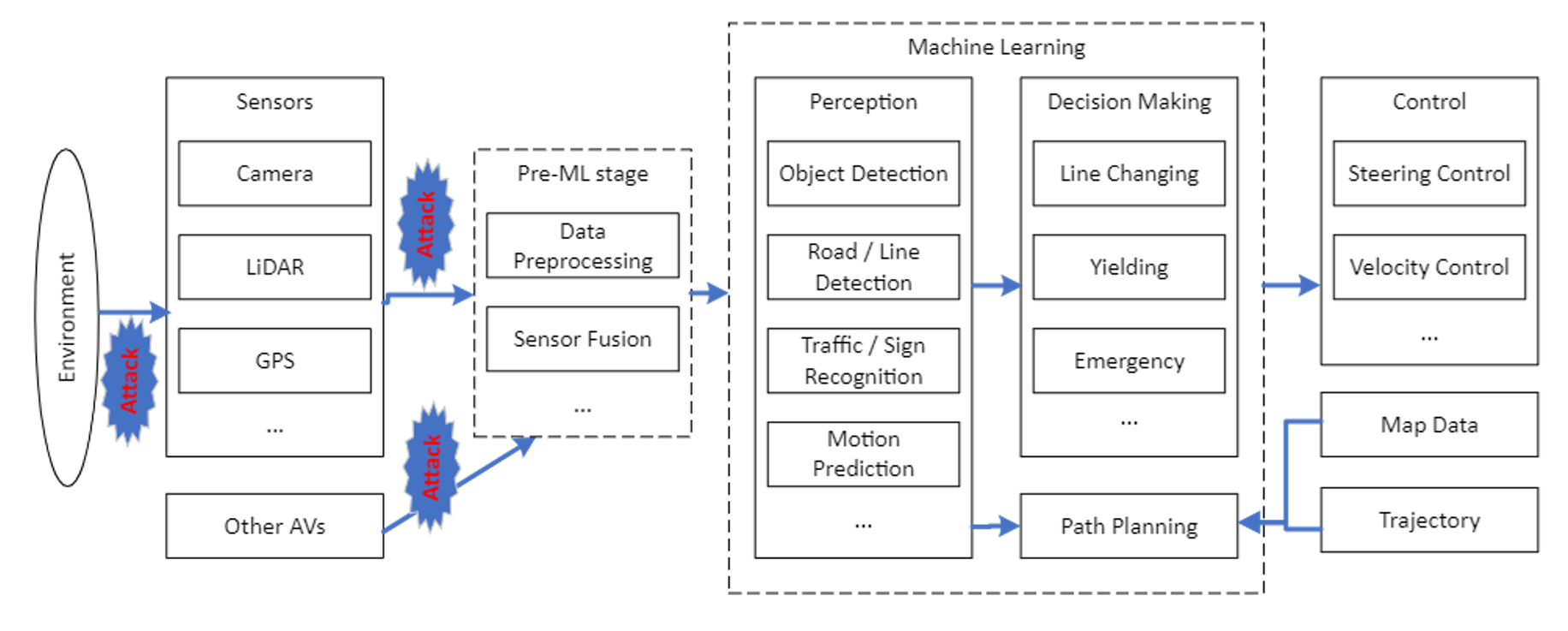}
	\caption{A Pipeline of Autonomous Vehicle System}
	\label{fig_2}
\end{figure*}

As we progress beyond Level 2 automation, the reliance on sensors like LiDAR (Light Detection and Ranging) radar, and cameras, coupled with machine learning algorithms becomes crucial~\cite{Marti2019}. Specifically, LiDAR's capability for precise distance measurements and high-resolution 3D mapping capabilities plays a vital role in improving environmental perception and situational awareness~\cite{Wandinger2005,McManamon2012,Diaz2017,Li2020,Benedek2021}. Over the last decade, LiDAR has become the most popular AV sensor as it enhances various aspects of autonomous systems, including obstacle detection, mapping, localization, and object recognition. Moreover, LiDAR-based autonomous systems are expected to have a significant impact on the Defence industry, promising efficiency and security in challenging terrains. Figure~\ref{fig_2} illustrates the pipeline of AV systems with sensors and the ML system, and potential points of attack. 

However, like any technology, ML-based LiDAR systems are not immune to vulnerabilities. Various attacks are possible, and our survey paper first concentrates on adversarial attacks, specifically traditional AML attacks and certain cyber-attacks on sensors that can cause malfunctions in ML-based perception modules. Attackers can employ AML techniques or cyber-attack methods to deceive or manipulate the perception systems of autonomous devices. These attacks can disrupt sensor inputs, potentially resulting in incorrect decisions and compromising safety and security. Subsequently, we proceed to investigate the theoretical aspects of defensive techniques against these adversarial attacks and analyze their limitations.

This survey paper addresses the gap in the existing literature by providing a comprehensive overview of ML approaches, adversarial attacks, and defenses specifically applicable to LiDAR-based systems in the context of autonomous driving. While some existing survey papers explored ML in autonomous driving~\cite{Liu2019a,Guo2020,Alaba2022,Zimmer2022}, they have primarily concentrated on ML functionality from a performance perspective, neglecting the importance of ensuring the robustness of the ML system against AML attacks or cyber-attacks on sensors targeting ML systems. In addition, unlike previous surveys that often focus on limited aspects of AML attacks, such as image spaces or specific attack methods~\cite{Bouchouia2023} and/or defenses~\cite{Guo2020} without in-depth analysis of their limitations, our survey takes a broader approach, encompassing the diverse spectrum of adversary threats and defenses in LiDAR-based perception systems. Moreover, our survey places a significant emphasis on identifying and analyzing the limitations of existing AML defense strategies in this domain. To achieve this, we begin with a literature review, examining 3D LiDAR sensory data and Machine Learning (ML) approaches in autonomous driving. By analyzing the current state of research on adversarial attacks, we aim to identify research gaps, emerging trends, and advancements in securing autonomous driving systems against adversarial threats.

Our survey aims to make significant contributions to the field:
\begin{itemize}
	\item {A Survey of 3D LiDAR-Based Machine Learning Models: Our survey paper investigates 3D LiDAR-based machine learning models in the context of autonomous driving, with a specific emphasis on widely recognized and highly cited models. By summarizing these prominent models in existing research, we provide readers with a clear overview of the state-of-the-art in this crucial area.}
	\item {Analyzing Adversarial Attacks and Defensive Strategies in 3D Autonomous Driving Systems: A central objective of our survey paper is to shed light on the evolving threat landscape facing autonomous driving systems. AML attacks and cyber-attacks present a significant challenge, potentially compromising the safety and security of these systems. Consequently, we conduct a comprehensive analysis of existing AML attacks and cyber-attacks tailored to the realm of 3D autonomous driving, as well as explore and analyze defensive strategies to mitigate these threats.}
\end{itemize}

Our survey covers a range of attacks, including cyber-attacks on sensors such as sensor spoofing, as well as physical attacks, and adversarial perturbations. We not only explain the methodologies behind these attacks but also explore potential defensive strategies and countermeasures. It is worth noting that existing defensive strategies often fall short of providing effective countermeasures. This paper highlights significant gaps in current research on autonomous system resilience against adversarial threats.

\section{Autonomous Vehicle (AV) System}\label{2AVSystem}
The processing pipeline of an AV system, depicted in Figure~\ref{fig_2}, comprises several stages that work together to enable the vehicle to operate autonomously. This pipeline includes sensors, preprocessing such as data pre-processing and sensor fusion, and ML modules like perception, decision-making, path planning, and control systems.

Sensors, such as cameras, LiDAR, GPS devices, etc., collect raw data from the environment. These sensors capture and relay information about the AV's surroundings in which the AV operates, including road infrastructure, traffic dynamics, environmental conditions, etc. for the AV's process. In this survey, we focus on AV systems using LiDAR sensors since LiDAR sensors offer a significant advantage over other sensors, such as radars and cameras, due to their higher resolution and precision. Additionally, LiDAR proves to be versatile, performing reliably under both daytime and night-time conditions~\cite{Luo2017,Javanmardi2019}.

Sensor data including LiDAR point cloud data requires pre-processing before being supplied to the ML algorithms. Data preprocessing plays a crucial role in preparing and refining the raw sensor data for further analysis. It involves data cleaning, filtering, and feature extraction, ensuring that the data input to the subsequent ML stages is of high quality and relevance. Sensor fusion, another vital part of pre-processing, combines data from multiple sensors to enhance perception accuracy. This fusion process is not necessary if the system has a single sensor. 

The ML component of the pipeline interprets and extracts meaningful insights from the pre-processed and fused sensor data. This step involves a range of tasks, including perception, decision-making, and route planning. The ML algorithms are central to providing an autonomous driving capability.

The perception module is responsible for sensing and comprehending the vehicle's environment. It processes data from an array of sensors, including LiDAR, cameras, GPS, and others, enabling the recognition of objects, identification of road features, and a holistic understanding of the environment. ML algorithms, specifically designed for object detection, semantic segmentation, object tracking, and scene comprehension, are integrated within this module. They work together to interpret sensor data, extract meaningful information, and create a representation of the environment that the AV can use for navigation. 

Once the perception system has gathered and processed sensor data, the next key step is decision-making. This module is dedicated to making important decisions on how to interact with the environment, based on the perceived environment. Here, the ML algorithm determines the vehicle's actions, including lane changes, yielding to other road users, and managing emergency scenarios. It heavily relies on the insights gained from the perception and ML components.

Path planning forms another integral part of the pipeline. It leverages map data and desired trajectory information to chart the optimal route for the AV. This component guides how the AV navigates from its current location to its destination, ensuring safe and efficient travel.

Finally, the control system executes the actions defined by the decision-making and path-planning stages. It manages the physical aspects of the AV, including steering control and velocity management, to enact the planned route accurately.

Throughout this intricate pipeline, there are various potential points of weakness, primarily at the interfaces between components and in communication with external entities. The ML perception typically handles the sensor data directly and processes it to understand the vehicle's surroundings. On the other hand, ML decision-making or planning does not typically deal directly with sensor data but relies on the processed information from the perception module to make decisions about how the vehicle should act. Thus, from an adversarial attack perspective, the ML perception module is exposed to potential attacks, as it directly interfaces with sensor inputs. Adversarial attacks on the perception module can manipulate the way the vehicle perceives its environment, potentially causing it to make incorrect or dangerous decisions. The ML decision-making module or planning module relies on the accuracy of the perception data, so it can indirectly be affected by attacks on the perception module. We describe ML-targeted attacks in Section~\ref{32Attacks}. In addition, AVs often communicate with other vehicles and infrastructure through communication systems. These communication channels can be vulnerable to attacks such as typical cyber-attacks or cyber-attacks on sensors described in Section~\ref{31Attacks}. Adversaries can send false data, manipulate data, or disrupt legitimate communication between vehicles, potentially causing accidents or traffic congestion. Therefore, robust security measures and vigilant oversight over the AV system are essential for protecting the AV against potential adversarial threats, ensuring the reliability and safety of autonomous driving technology. 

In this survey, our primary focus is the robustness of LiDAR-based autonomous driving systems. The LiDAR system generates data that reveals the presence of obstacles in the environment and the vehicle's relative position to these obstacles. This data offers insights into the contours of roads, nearby infrastructure, and vegetation. In the following subsection, we investigate the LiDAR system, the characteristics of LiDAR data, and the machine learning approaches applied to analyze this data. We then provide a literature survey regarding the ML perception modules that are related to the robustness of AV systems. 

\subsection{LiDAR System}\label{21LiDARSystem}
In a typical LiDAR system, a laser emits infrared light pulses at various angles onto target objects. The LiDAR sensor subsequently detects and processes the reflected echoes, creating a 3D point cloud. Each point in the cloud is represented by a 3-dimensional vector (x, y, z), providing spatial coordinates within the LiDAR coordinate system. Additional features like color or intensity can be integrated if needed. This point cloud is a fundamental element that can be further processed and analyzed for a wide array of applications, including environment and object perception, localization, and object recognition.

LiDAR, however, has limitations. It requires a direct line of sight of objects and may face challenges in environments with obstacles or occlusions~\cite{Samman2000,Li2020}. A LiDAR can only see objects that reflect its signal. If the signal does not return, whether due to absorption, transparent materials, or reaching its range limits, LiDAR interprets it as an absence of objects. Reflective surfaces may introduce data inaccuracies or false readings in LiDAR measurements. Furthermore, adverse weather conditions such as heavy rain, fog, or snow can disrupt LiDAR data collection, diminishing its reliability under unfavorable weather conditions. 

The inherent nature of LiDAR point cloud data is characterized by its sparsity and lack of structure, rendering the processing of LiDAR data a challenging endeavor. Primary tasks associated with point cloud data encompass 3D object detection, semantic/instance segmentation, motion prediction, multiple object tracking, localization, object detection and classification, scene segmentation, and scene understanding. Object detection systems, exemplified by PointPillars~\cite{Lang2019} , VoxelNet~\cite{Zhou2018}, or SECOND~\cite{Yan2018}, operate by taking a 3D point cloud as input and subsequently generating bounding boxes around objects within the environment. This functionality provides spatial awareness and enables AVs to make informed navigation and control decisions.

The processing of LiDAR point clouds necessitates the deployment of state-of-the-art models that heavily utilize Deep Neural Network (DNN) architectures. These DNNs, while powerful, are susceptible to adversarial attacks. This vulnerability is particularly pronounced in the perception component of ML models utilized in autonomous driving. Sensitivity to input variations makes these models vulnerable to adversarial perturbations, compromising their ability to accurately detect, segment, or map objects. These incorrect perceptions can then lead to erroneous decisions in the downstream decision-making component of the autonomous system. For example, the injection of a false object may lead to ML decision to apply emergency brake operations that may injure passengers, resulting in severe safety consequences.

\subsection{LiDAR Point Cloud}\label{22LiDARPointCloud}
Three-dimensional LiDAR provides high-resolution point clouds with accurate representations of the length, width, and height of objects, offering a comprehensive view of the environment. However, 3D LiDAR generates a substantial amount of data per scan due to its detailed output. The extraction and interpretation of geometric details from 3D range data are notably more intricate in comparison to 2D data. Moreover, 3D laser data introduces an additional complication as the lower layers of the scanner frequently capture ground or floor surfaces, further adding to the complexity of data analysis.

Reliable perception of the surrounding environment is typically achieved through two subtasks: simultaneous localization and mapping (SLAM) and detection and tracking of moving objects (DATMO)~\cite{Azim2012}. SLAM aims to create a map comprising static elements of the environment, providing the vehicle with knowledge of its surroundings. On the other hand, DATMO utilizes this map to detect and track dynamic objects in real-time.

Point clouds are characterized by several properties that make them challenging to work with:
\begin{itemize}
\item{Massive: Point clouds can be massive, especially when representing complex scenes or large environments. Each point in the point cloud requires storage for its 3D coordinates and potentially additional features like color or intensity. }
\item{Noise: Point clouds often contain noise, which refers to unwanted or random variations in the measured point coordinates or features. Noise can be caused by numerous factors, such as sensor inaccuracies, environmental conditions, or reflections. }
\item{Incomplete: In real-world scenarios, it is challenging to obtain complete and perfectly sampled point clouds. Incomplete point clouds arise when some parts of the 3D scene are occluded, not captured by the sensor, or simply missing due to limitations in data acquisition. }
\item{Irregular: Point clouds are considered irregular because there is no predefined structure or grid-like organization to the points. Unlike images, which have a fixed grid of pixels, points in a point cloud can be distributed arbitrarily in 3D space.}
\item{Unstructured: Point clouds lack inherent connectivity or spatial relationships between points. Each point is independent, with no knowledge of its neighbors or their arrangement. They are collected from laser reflections off surfaces, without a predefined pattern like images.}
\item{Unordered: The order of points in a point cloud is arbitrary, and there is no predefined sequence or arrangement. Different scans or data acquisition processes can result in different orderings of the points in the point cloud. }
\item{Sparse: Although LiDAR provides accurate distance measurements, the point clouds are sparse and have non-uniform densities across the scene. This sparsity is a result of factors such as the sensor's range, resolution settings, and the geometry of the environment being scanned. It means that there are fewer points available to accurately capture the nuances of the environment, making it harder for ML algorithms to accurately understand and navigate their surroundings.}
\end{itemize}

These properties make point cloud processing a challenge in the field of computer vision and 3D data analysis. Specialized algorithms and deep learning architectures, such as PointNet~\cite{Qi2017a}, have been developed to handle these properties and extract meaningful information.

\subsection{Machine Learning for LiDAR Data}\label{23ML}
The primary focus of this paper is on the ML perception module, which is considered a central point of vulnerability due to its pivotal role in comprehending the environment. While decision-making is also a significant concern, it typically relies on the information processed by the perception module to guide the vehicle's actions. Furthermore, communication vulnerabilities, while important, may have different implications, as they can impact the ML perception modules. Therefore, we prioritize our focus on ML perception, as it is directly impacted by adversarial threats.

As introduced earlier, 3D point cloud data holds a significant role in autonomous driving, serving as a vital component within the perception systems of self-driving vehicles. Processing LiDAR data can be challenging due to the characteristics of point clouds outlined in Section~\ref{22LiDARPointCloud}. Traditional deep learning techniques, such as Convolutional Neural Networks (CNNs), have primarily been designed to work with data organized on structured grids, like 2D images (composed of pixels) or 3D volumes (represented as voxels). In such grid-based structures, adjacent data points have well-defined relationships, enabling efficient application of convolutional operations. In contrast, point clouds possess inherent structural irregularity. They consist of individual points scattered throughout 3D space, lacking any predefined grid or regular arrangement. Notably, LiDAR-generated point clouds are less susceptible to adversarial attacks compared to images~\cite{Xiang2019,Zhu2021,Tu2021}, making it more challenging to launch attacks on LiDAR-based AVs. Nevertheless, it is important to recognize that, while challenging, such attacks are not impossible and there has been a recent increase in their prevalence and sophistication as discussed in Section~\ref{3Attacks}.

There have been several survey papers on deep learning approaches using 3D LiDAR data~\cite{Liu2019a,Bello2020,Guo2020,Alaba2022,Zimmer2022}. In this section, we highlight the most popular approaches commonly targeted for AML attacks. We explore a range of popular deep learning models that differ in their approaches, especially regarding feature extraction, which is a principal factor in processing 3D point clouds.

PointNet~\cite{Qi2017a} is a pioneering deep learning model that addresses the challenge of processing irregular and unstructured 3D point cloud data. PointNet's main goal is to extract meaningful features and patterns from these point sets without relying on any predefined order or connectivity. It achieves this by utilizing continuous symmetric functions, approximated through shared Multi-Layer Perceptrons (MLPs), to process the 3D coordinates and additional features, such as colors, of each point independently. As a result, PointNet exhibits permutation invariance, enabling it to handle point clouds regardless of the order in which points are presented. Additionally, PointNet generates embeddings for each point within a point cloud, and these embeddings undergo transformation and aggregation to ensure invariance to geometric transformations, such as rotations or translations. This property ensures that the classification results remain unaffected by rotations of the input point clouds.

PointNet++~\cite{Qi2017b} is an extension of PointNet that introduces a hierarchical feature learning approach. It organizes the points into nested sets and processes them hierarchically, enabling the model to capture both local patterns and global context in the point cloud data. PointNet++ is specifically designed for point set segmentation tasks. It achieves this by subsampling the point cloud into overlapping regions (query points) and processing each region using a 'Set Abstraction Layer' to extract local features. These features are then interpolated and propagated back to the original points to capture global contextual information.

PointNet and PointNet++ serve as foundational architectures for processing raw 3D point cloud data directly. PointNet-based models are commonly employed in tasks such as object detection, segmentation, and scene understanding. Frustum PointNet~\cite{Qi2018} extends the PointNet architecture to fuse LiDAR and camera data for increased perception reliability. The Frustum PointNet approach begins by creating 3D bounding boxes called frustums around objects initially detected in 2D images. These frustums define regions of interest within the 3D point cloud data. Subsequently, PointNet is applied to process the point cloud data within these frustums, enabling tasks such as object detection and localization in 3D space. Similar to Frustum PointNet, Frustum ConvNet~\cite{Wang2019a} uses PointNet operations at lower layers of its network architecture. However, Frustum ConvNet employs convolutional layers as part of its architecture, for capturing spatial relationships and aggregating local features, whereas Frustum PointNet relies on fully connected layers. At these lower layers, Frustum ConvNet may share similarities with Frustum PointNet in terms of processing individual points within frustums using shared MLPs. Frustum ConvNet extends beyond this by incorporating convolutional layers, which enable it to capture spatial relationships and context more effectively.

Voxel grids are used to convert point clouds into a 3D grid format, making it easier to apply 3D convolutions. VoxelNet~\cite{Zhou2018} is a deep learning model that leverages 3D voxel grids to represent 3D data. It introduces a 3D detection network tailored for precise object detection in sparse LiDAR point clouds. The approach involves partitioning the point cloud into uniformly spaced 3D voxels, a random selection of a fixed number of points from each voxel, and utilizing a voxel feature encoding layer to construct a comprehensive volumetric representation. VoxelNet unifies the tasks of feature extraction and bounding box prediction within a single, end-to-end trainable deep network, eliminating the necessity for manual feature engineering. In 3D object detection workflows, PointNet can be utilized as a feature extractor to process individual points within the voxels generated by VoxelNet. This facilitates more detailed feature extraction at the point level within each voxel. It is important to note that PointNet is not an intrinsic component of the VoxelNet architecture; its integration depends on the specific implementation. Furthermore, VoxelNet organizes point cloud data into a 3D voxel grid, effectively establishing a Bird's-Eye View (BEV) representation, which is particularly advantageous for object detection tasks.

SECOND~\cite{Yan2018} introduced sparsity-aware convolutional layers, which make it efficient in handling sparse LiDAR data without the need for voxelization. LiDAR sensors often produce sparse point cloud data, meaning that some regions may have no data points at all, and others may have varying point densities. SECOND employs sparsity-aware convolutional layers, which are specialized for processing sparse data efficiently. Instead of applying convolution operations to all points uniformly, they adaptively select and process only the relevant points, effectively ignoring empty regions with no points. In addition, when processing a point in a sparse region, the sparsity-aware layers consider their local neighborhood of points. This neighborhood may vary in size and shape depending on the point distribution. The layers extract features from this dynamic neighborhood.

PointPillars~\cite{Lang2019} initially transforms the original point cloud data into a top-down, 2D BEV representation of the 3D environment. This 2D perspective is further analyzed and feature extraction is performed using 2D convolutional layers. The BEV representation is then subdivided into a grid composed of "pillar-like structures". These pillar-like structures are essentially the grid cells used to organize the 3D point cloud data in the BEV representation. Within each of these pillars, PointPillars utilizes 2D convolutional layers to process the points and extract features. PointPillars operates in real-time for object detection, allowing it to predict object candidates belonging to multiple classes. It provides estimations of their 3D-oriented bounding boxes and associated class confidence values. 

PIXOR~\cite{Yang2018a} is another 3D object detection architecture that also has its feature extraction approach tailored to the task. PIXOR operates on 2D camera images, specifically focusing on LiDAR-camera fusion for 3D object detection from BEV. The model utilizes a 3D occupancy grid representation with accumulated reflectance and employs a classification heat map and regression features for object localization. The BEV representation is preferred due to its accuracy and avoidance of object overlap, enhancing computational efficiency. The top-down perspective in BEV eliminates depth ambiguity ensuring clear separation of objects at different distances. PIXOR is one of the fastest LiDAR object detection models and is further improved in PIXOR++~\cite{Yang2018b}.

PointRCNN~\cite{Shi2019} is an approach in 3D object detection compared to previously mentioned methods like PointNet, VoxelNet, PointPillars, and PIXOR, in terms of its feature extraction and overall approach. Unlike traditional CNNs designed for structured grids, PointRCNN operates directly on unstructured point cloud data, allowing it to capture intricate spatial information for precise object detection in 3D scenes. PointRCNN introduces a two-stage detection framework that first generates region proposals using a ‘Region Proposal Network’ and then refines these proposals using 3D CNNs. The proposals are often represented in a BEV for object detection. Both PointRCNN and PointNet are capable of directly handling unstructured 3D point cloud data, but PointRCNN is specifically designed for accurate 3D object detection and localization tasks, while PointNet is for a broader range of tasks beyond object detection, including segmentation, classification, and scene understanding due to the strengths of its permutation and transformation invariance.

MotionNet~\cite{Wu2020} presents a distinctive approach to motion prediction using 3D point cloud data. It achieves this by converting 3D point clouds into BEV maps, streamlining subsequent computations. The model encompasses various features, including the ability to generalize to unseen objects, integration of temporal information, the capture of multi-scale spatio-temporal features, etc. MotionNet's architecture relies on standard 2D and pseudo-1D convolutions, rendering it well-suited for real-time operations in autonomous driving scenarios. However, its performance may exhibit variability based on the specific application.

Some studies focus on the fusion of multiple sensors with LiDAR. Ku et al.~\cite{Ku2018} introduce an approach for 3D object detection that can jointly generate 3D object proposals and perform object detection by aggregating information from multiple views or camera angles. This multi-view approach is designed to enhance the accuracy and robustness of 3D object detection, especially in situations where objects might be partially hidden or obstructed when viewed from a single perspective. EPNet~\cite{Huang2020a} is also a system that combines information from 3D data sources with image data to enhance object detection capabilities. It integrates image semantics to provide additional context and information, which in turn improves the accuracy of object detection.

In addition to the above-mentioned approaches, various other deep learning approaches, including those proposed by Huang et al.~\cite{Huang2020b} and Priya and Pankaj~\cite{Priya2021}, have been developed to process 3D LiDAR point cloud data. Given the inherent vulnerabilities in deep learning, it is necessary to prioritize efforts aimed at identifying relevant adversarial attacks and defenses. In the following section, we will explore adversarial attacks.

\section{Attacks}\label{3Attacks}
In the realm of computer science, an ‘adversary’ refers to those attempting unauthorized access or corruption of a network. Originating in 2004 for anti-spam filter robustness~\cite{Dalvi2004}, adversarial machine learning has evolved to challenge the security of ML models. Autonomous vehicles, heavily reliant on a diverse network of sensors, including LiDAR, radar, cameras, and GPS, leverage advanced ML algorithms for processing multi-modal inputs. These algorithms play a pivotal role in environment perception and vital operational decisions. While sensors are traditionally deemed trusted components in AV control systems, any compromise to their integrity, resulting in falsified readings, introduces significant risks to both vehicle safety and security. These risks stem from a two-fold vulnerability: both sensors and ML algorithms are susceptible to adversarial attacks. 
\begin{itemize}
\item{Sensors: The sensors themselves are susceptible to a range of conventional cyber-attacks, including authentication breaches, Denial-of-Service (DoS), jamming attempts, or even direct physical attacks~\cite{Teixeira2012,Petit2014,Sanchez2019,Manvi2017}. Alongside these established threats, malicious actors can exploit these vulnerabilities by targeting the AV communication channel to execute various deception attacks, such as Sybil, spoofing, or replay attacks as elaborated in Section~\ref{31Attacks}. These deceptive tactics can result in the manipulation or corruption of sensor data, leading to erroneous perceptions and decisions by the autonomous vehicle. }
\item{ML Algorithms: Beyond sensor vulnerabilities, AVs face a distinct susceptibility to adversarial targeting in the realm of ML models. Attackers can exploit weaknesses in these algorithms, deceiving vehicles into making unsafe decisions. Such manipulations may include subtle alterations to the training data, injection of backdoor triggers, or the introduction of carefully crafted adversarial perturbations into the sensory inputs, challenging the ML model’s recognition capabilities.  Notably, these perturbations do not alter the semantic meaning of the scene but wield a transformative influence on the ML model’s output. Section~\ref{32Attacks} provides a comprehensive exploration of adversarial attacks on ML algorithms.}
\end{itemize}
In the context of LiDAR based AVs, attacks by adversaries encompass a range of tactics. These tactics include adversarial manipulation in the image domain, along with attacks involving manipulation of sensor inputs, introduction of fabricated data, or repetition of attacks, among other strategies. The overarching objective is to exploit weaknesses in the underlying ML algorithms, ultimately deceiving the vehicle's perception and potentially influencing its driving decisions for malicious purposes. 

\subsection{Attacks on Sensors}\label{31Attacks}
LiDAR sensor systems are susceptible to attacks that can compromise their integrity, availability, and accuracy~\cite{El-Rewini2020,Giannaros2023,Girdhar2023}. These attacks intend to mislead autonomous systems by altering data from LiDAR sensors, which can result in incorrect perceptions and unsafe decisions by autonomous vehicles.
 
Attackers can accomplish this by emitting deceptive signals to manipulate distance measurements or introducing false objects to the LiDAR perception system. We focus on the cyber-attacks aiming to disrupt or manipulate the LiDAR readings which eventually lead to disrupting the ML decision making. Since sensors are typically considered trusted components in an AV’s control system, falsified readings could lead to unforeseen consequences if the sensors are compromised. 

Sensor attacks typically occur within the vehicle’s communication channel and have the potential to manipulate LiDAR sensor data in several ways, such as creating fake objects (Sybil attacks), injecting malicious points in LiDAR data (Spoofing attacks), or even replaying outdated point clouds (Replay attacks) to deceive the AV system. 

\subsubsection{Spoofing Attacks}\label{311Spoofing}
Spoofing, a form of masquerading attack, severely impacts the trustworthiness of LiDAR systems. In spoofing attacks, a spoofing device emits laser pulses toward the victim LiDAR, which disrupts the timing of the laser reception events and consequently, alters the calculated 3D positions of objects. This manipulation results in the creation of ‘spoofed points’ within the LiDAR's point cloud. These spoofed points can lead to object misdetection by the downstream object detector. For instance, the attack can relocate points originally associated with an object~\cite{Petit2015} or even render the object undetectable~\cite{Cao2023}. Alternatively, spoofed points clustered together can falsely trigger the detection of a non-existent object~\cite{Hau2021}. Spoofing attacks can be synchronized or asynchronized. The synchronized attacks require precise knowledge of the victim LiDAR’s scanning pattern beforehand to synchronize the malicious laser firing timing~\cite{Cao2019a,Sun2020a}. Asynchronized attacks do not need such knowledge and thus are more deployable~\cite{Petit2015,Shin2017}. 

Petit et al.~\cite{Petit2015} highlighted how attackers can manipulate a LiDAR system through an asynchronous attack, relaying laser signals from a different location to generate fake data points that appear farther away than their actual positions. Building on this, Shin et al.~\cite{Shin2017} further illustrated the potential to inject false data points, creating illusions that seem closer to the location of a spoofed device. These attacks distort perception and can disable the LiDAR's ability to sense specific directions. The authors proposed a blinding attack wherein the LiDAR is exposed to an intense light source with the same wavelength as the LiDAR. As a result, the LiDAR failed to perceive objects from the direction of the light source. Additionally, Park et al.~\cite{Park2016} demonstrated spoofing attacks on infrared sensors used in medical infusion pumps, resulting in the manipulation of medicine dosage. 

Sato et al.~\cite{Sato2023} demonstrated an improved LiDAR spoofing capability tailored for contemporary LiDAR systems. The key idea is to fire many attack laser pulses at a higher frequency than the victim LiDAR's laser-firing frequency to achieve the spoofing effect for every point in the scanning range. The attack's effectiveness depends on the high frequency of the attack laser pulses.

\subsubsection{Replay Attacks}\label{312Replay}
Replay attacks have been extensively studied in the fields of estimation and control ~\cite{Mo2009,Teixeira2012,Porter2021}. In a replay attack, a message is recorded and subsequently played back at a different time with malicious intent.  The specific objective in the context of LiDAR systems is to mislead the perception system, causing a miscalculation of the target's time and position~\cite{Petit2015,Hallyburton2023}. For instance, a vehicle in front of the victim AV is moving forward at a speed of 110 km/h. The attacker captures the LiDAR data from the victim AV and stores it for later use. When the leading vehicle slows down, the adversary injects the old LiDAR data into the system periodically. The victim AV still believes that the leading vehicle is traveling at 110 km/h and may not decelerate appropriately, potentially causing an accident.

In a study by Stottelaar~\cite{Stottelaar2015}, a LiDAR replay attack was demonstrated wherein signals were recorded for later insertion of malware, which could lead to false detections of non-existent objects. Consequently, the vehicle's control unit might erroneously perceive a significant obstacle and initiate an abrupt stop. Similarly, Petit et al.~\cite{Petit2015} employed a method where they captured, delayed, and then replayed the LiDAR signal, causing the LiDAR sensor to produce inaccurate measurements.

\subsubsection{Sybil Attacks}\label{313Sybil}
Traditionally employed in peer-to-peer networks, sensor networks, and vehicular ad hoc networks (VANET), Sybil attacks are now being extended for multimodal interconnected sensors and communication technologies within AV systems~\cite{Sinai2014,Nanda2019,Cui2019,Bouchouia2023}. The term ‘Sybil attack’ originates from the book Sybil, which narrates the experiences of an individual diagnosed with dissociative identity disorder~\cite{Douceur2002}. In a Sybil attack, adversaries exploit the vehicle communication infrastructure to generate and advertise multiple fake identities (Sybil nodes), disrupting the environmental perception of victim AVs~\cite{Douceur2002}. This falsified information attack can be orchestrated through a malicious vehicle or without the involvement of an actual vehicle. The fake identities, whether stolen from inactive nodes or newly fabricated, can pose as non-existing objects or fake vehicles, violating the assumption of accurate traffic perception by AV sensors~\cite{Vasudeva2018}. This subversion compromises object detection and tracking, leading to navigation errors and potential traffic chaos, particularly among LiDAR-based AVs~\cite{Rabieh2015}. For instance, an attacker can use Sybil nodes to issue false warnings, creating a fabricated traffic jam or road incidents, influencing victim AVs to alter their routes, while false object signatures exacerbate confusion for the LiDAR sensors. More advanced Sybil attacks can involve spoofing other sensors like GPS and LiDAR, potentially leading to disastrous traffic incidents~\cite{Shoukry2018}. Furthermore, Sybil nodes can disrupt LiDAR sensors via DoS attacks, overloading communication channels and compromising accurate surrounding detection and mapping. 

LiDAR generates a point cloud frame of its surroundings, offering data on nearby objects, such as their distances and angles from the vehicle. In context of Sybil attacks, if fake objects are introduced into the surroundings, LiDAR will feed this misinformation into the backend ML algorithm, leading to confused or wrong perceptions of the environment. This misinformation may trigger the AV system to indicate traffic congestion, resulting in unnecessary slowdowns. In interconnected vehicle environments, it may force other vehicles to change their routing decisions, avoiding falsely congested or targeted areas. Discrepancies may arise if data from other sensors, such as GPS, radar, and cameras, do not align with the LiDAR-generated information. Lim et al.~\cite{Lim2020} conducted experiments on Sybil attacks in VANETs using multiple sensor sources, including LiDAR. They also proposed defenses as discussed in Section~\ref{41Defenses}. 

It is important to note that Velodyne VLP-16 has been the standard choice for LiDAR robustness evaluation in most research. Studies have either exclusively assessed their attacks on the VLP-16 or used its attack susceptibility to validate their threat models. Sato et al.~\cite{Sato2023} have argued through their experiments that sensor attacks relevant to  VLP-16 may not necessarily apply to more recent LiDAR systems referred to as "next-generation" LiDARs~\cite{Yoshioka2022}. The next-generation LiDARs incorporate advanced security features, including laser timing randomization and pulse fingerprinting. While prior works~\cite{Shin2017,Cao2019a,Cao2023} have mentioned these features as potential defenses, none have thoroughly assessed their effectiveness against Sybil, spoofing, or replay attacks. By randomizing the timing of laser pulses, it becomes exceedingly difficult for an attacker to predict when a pulse will be emitted. This unpredictability disrupts the attacker's ability to synchronize their malicious signals with the LiDAR system's emissions, rendering pre-calibrated attack timings obsolete. Interestingly, we find that the attacker may still be able to inject some points if the randomization is not strong enough. Pulse fingerprinting assigns unique signatures to laser pulses, enabling the LiDAR system to authenticate data and prevent replay attacks. Digital signatures also offer authentication and non-repudiation for laser pulses. Using a nonce in laser beacons, generated uniquely each time, thwarts replay attacks by making intercepted data unusable in subsequent communications.

\subsection{Attacks on ML}\label{32Attacks}
This section explores studies on attacks targeting ML models behind LiDAR-based perception and decision-making processes, aiming to understand the strategies employed by malicious actors and potential countermeasures in Section~\ref{4Defenses}. We also delve into the complex landscape of multi-modal attacks, which introduce a new layer of security challenges for autonomous vehicles equipped with diverse sensor modalities. 

Since the publication of Szegedy et al.~\cite{Szegedy2013} in 2013, adversarial attacks have attracted considerable attention from both research and the hacker communities. These attacks involve deliberate data manipulations to fool an ML model’s decision-making process. They can take various forms depending on the adversary’s goals and capabilities, including  evasion attacks during inference~\cite{Cao2019b}, poisoning attacks to corrupt model training~\cite{Xiang2021}, or model extraction~\cite{Zhang2022a}. Depending on the attacker’s knowledge, attacks can be categorized as white box~\cite{Cao2019a,Goodfellow2014} or black box~\cite{Papernot2017,Redmon2016}. In white box attacks, the adversary is assumed to know the target model’s internal architecture, training data, and parameters, while in black box attacks, the attacker can only query the model outputs. 

The vulnerability of LiDAR systems to adversarial attacks is an emerging concern. Despite advances in LiDAR-based computer vision and environment perception, the underlying ML algorithms, which are primarily based on DNNs, are vulnerable to carefully crafted adversarial examples. These examples are subtle modifications to input data that are imperceptible to humans but significantly alter the model's output, leading to incorrect predictions or classifications. The objective of such attacks is to manipulate LiDAR point clouds discreetly, deceiving the ML classifier without detection.

Attacking specialized LiDAR-based object detection model is more complex than attacking 2D object detection or image classification model~\cite{Xiang2019,Tu2021,Redmon2016}. This complexity arises from the need to deceive multiple region proposals, representing potential bounding boxes around objects in a 3D space. Unlike 2D object detection, where the goal is to precisely identify and locate objects within the image plane, LiDAR-based object detection requires capturing objects in a 3D space. Compared to image classification, where the objective is to assign a single label to the entire image, object detection with LiDAR involves the identification and localization of multiple objects in a 3D environment. Adversaries attempting to compromise object detection models for LiDAR must craft highly accurate adversarial perturbations. These perturbations must effectively mislead the model's class predictions and adjust the localization of objects within each proposed region. This level of precision demands more intricate and precise adjustments compared to adversarial attacks on 2D object detection and image classification models~\cite{Xiang2019,Zhang2024,Li2021a,Lehner2022}. LiDAR data consists of numerous 3D points, each with unique coordinates and intensity values, intrinsically different from 2D images. Determining how changes in individual points affect the system's decisions is a more intricate task compared to straightforward pixel-level alterations in images. For example, unlike image-based object detection where perturbations only alter pixel values, perturbing point clouds may alter associated voxels in voxel-based object detectors, resulting in inconsistencies in the environment representation~\cite{Yan2018}. Moreover, the irregular point cloud structure complicates the adaptation of existing adversarial attacks from the 2D image domain. Unlike 2D images, the flexibility to add points at any 3D position creates an extensive search space for generating adversarial examples~\cite{Xiang2019}. Additionally, models processing 3D point clouds have entirely different structures than 2D deep learning models, introducing unique properties that increase the complexity of adversarial attacks~\cite{Li2021a}.

Ongoing research explores semantic adversarial attacks, which concentrate on modifying specific attributes like lighting conditions and input clarity to create genuine adversarial instances~\cite{Deng2021}. This highlights that deep learning models can err in real-world scenarios even without external adversaries~\cite{Lehner2022,Hendrycks2021}. Factors such as weather and lighting conditions can inadvertently give rise to semantic adversarial attributes, presenting unforeseen risks to AVs. Hamdi and Ghanem~\cite{Hamdi2020} proposed a theoretical framework for analyzing DNNs' robustness in semantic space and introduced a method to detect robust regions within networks. Through extensive experiments, they reveal variations in semantic robustness among well-known network architectures, challenging assumptions about accuracy and robustness.

While LiDAR data exhibits resilience against specific forms of noise or perturbations, enhancing its robustness~\cite{Xiang2019,Zhu2021,Zhang2024}, adversarial attacks encompass a range of techniques often involving subtle alterations. Since LiDAR data may already contain some degree of noise due to sensor limitations, introducing significant perturbations without detection poses a challenge for attackers. The AVs collect consecutive frames of point cloud data, and altering just one frame could be interpreted as a mistake by the perception system. To succeed in their attack, attackers must change all consecutive frames, but predicting AVs' driving behavior is usually beyond their capability~\cite{Zhu2021}.

Next, we will discuss the categorization of ML attacks on LiDAR systems into three main types: Evasion attacks, Poisoning attacks, and model stealing attacks.

\subsubsection{Evasion Attacks}\label{321Evasion}
Evasion attacks target trained ML models during inference time~\cite{Jiang2020}. Attackers craft subtle alterations to input data to deceive the model's decision-making. The goal is to generate adversarial examples that can evade detection by the model, resulting in incorrect predictions. By attacking the underlying ML models of LiDAR systems, especially deep learning models, these attacks can compromise the AV’s ability to accurately perceive its environment and make informed decisions.

Xiang et al.~\cite{Xiang2019} pioneered an optimization algorithm based on the Carlini and Wagner framework~\cite{Carlini2017} for generating targeted attacks for misleading the PointNet model. Their method can craft adversarial examples by perturbing existing point cloud data or generating a new one. They tackled the exhaustive search space of 3D point cloud by initializing the points at salient locations and then optimizing their positions. Liu et al.~\cite{Liu2020} extended upon the Fast Gradient Sign Method (FGSM)~\cite{Goodfellow2014} to explore two categories of adversarial attacks: distributional attacks that involve imperceptible perturbations to the distribution of points, and shape attacks that involve deforming the shape represented by a point cloud. Their method constrained the perturbation magnitude onto the surface of an epsilon ball in different dimensions. Zhou et al.~\cite{Zhou2020} introduced LG-GAN, a label-guided adversarial network designed for real-time point cloud classification. By inputting the original point cloud and the target label, LG-GAN deforms the point cloud with a single forward pass, thereby assigning it an incorrect label.

In addition to shifting or adding new points to the original point cloud, some researchers have proposed techniques to generate adversarial point clouds by strategically removing points. Zheng et al.~\cite{Zheng2018} introduced a malicious point-dropping technique by learning a saliency map. A saliency map assigns scores to points in a point cloud data, reflecting their contribution to the model recognition loss. High scores highlight significant segments in the 3D scene. Yang et al.~\cite{Yang2019} proposed a point-detach strategy that iteratively detaches the most important point to confuse the attacked network, leveraging saliency maps. They also developed a variant of the one-pixel attack~\cite{Su2019}, employing a pointwise gradient method to update only the attached points without altering the original point cloud. Additionally, Wicker and Kwiatkowska~\cite{Wicker2019} investigated adversarial attacks on point cloud detection models, presenting an iterative saliency occlusion approach to generate adversarial point cloud examples by dropping key points. 

Zhao et al.~\cite{Zhao2020} highlighted the potential failure of most of the point cloud manipulations discussed above due to the isometric robustness of point cloud ML models. Isometry represents geometric transformation like rotation, translation, and scaling that preserves the geometric shape and size of the point cloud. Their findings revealed significant degradation in perception performance under adversarial rotations. Through transferable black-box attacks employing Thompson Sampling (TS)~\cite{Russo2018} and the Restricted Isometry Property (RIP) matrix~\cite{Candes2005}, their research uncovered vulnerabilities in 3D deep learning models. RIP quantifies deviation from isometry so that the model remains robust under slightly perturbed isometry, whereas TS is favored for its adaptability in learning successful attacks likelihood across different rotation ranges. 

Hu et al.~\cite{Hu2023} developed the PointCA adversarial attack directed at 3D point cloud completion models, which reconstruct incomplete point cloud data observed by AV perception modules. Adversarial attacks on these completion models aim to produce deceptive geometric shapes rather than incorrect class labels. PointCA generates adversarial point clouds that resemble the originals but are completed as different objects with altered semantic information. By employing neighborhood density information, their algorithm tailors perturbations to match local geometry, ensuring stealthy attacks across diverse point cloud distributions. 

In addition to these evasion attacks, two prominent strategies for evading the perception and decision-making systems of AVs are physical adversarial attacks and adversarial spoofing. Physical adversarial attacks involve directly tampering with the physical environment to mislead the LiDAR systems, for example, by placing physical objects or reflective surfaces strategically to create false readings. Adversarial spoofing involves strategically manipulating LiDAR data, to deceive the perception system into misclassifying objects or perceiving non-existent obstacles.

Tu et al.~\cite{Tu2020} devised a method to create physically realizable adversarial examples that can be placed on a vehicle and make this vehicle invisible to the LiDAR-based 3D object detectors. By placing a 3D-printed adversarial example object on top of a vehicle, the vehicle becomes partially invisible to the targeted LiDAR detector system. However, creating such precisely shaped objects with abnormal features is difficult to achieve with high precision in practice. Another attack LiDAR-Adv~\cite{Cao2019b} introduced a gradient optimization-based approach for generating a 3D printable physical adversarial object. This object can be placed on the road without being detected by LiDAR. Zhu et al.~\cite{Zhu2021} studied the vulnerability of LiDAR point cloud semantic segmentation and introduced a method to identify the adversarial locations in a 3D scene, so that arbitrary objects such as road signs and cardboard placed at these locations can fool the LiDAR perception systems. Cao et al.~\cite{Cao2021} propose to generate a physically realizable and adversarial 3D object that is invisible to both the camera and LiDAR sensor. It focuses on the impact of adversarial physical objects on camera-LiDAR perception. The adversarial object is generated through optimization and can be leveraged to attack the multi-sensor fusion-based 3D object detection models. MSF-ADV first generates 3D objects of different shapes so that they can simultaneously affect the LiDAR-based 3D point cloud imaging and the RGB color values of the pixels in the RGB image. Secondly, MSF-ADV uses an optimization algorithm to generate the 3D shapes with the best adversarial effect. Finally, MSF-ADV uses a 3D printer for the physical generation of adversarial objects. 

Cao et al.~\cite{Cao2019a} were the first to strategically inject spoofed LiDAR points in a white-box optimization-based attack, deceiving an AV into detecting a non-existent vehicle during post-processing. They simulated on-road scenarios by introducing carefully crafted adversarial laser points into the initial 3D point cloud obtained by the Baidu Apollo team, using the Velodyne VLP-16 LiDAR sensor. Their research focused on AML techniques for manipulating LiDAR perception, presenting emergency brake and AV freezing attack scenarios. Their experiments showed success rates of up to 90\% when injecting over 60 adversarial points. Sun et al.~\cite{Sun2020a} exploited overlooked occlusion patterns in LiDAR point clouds, making vehicles susceptible to spoofing attacks. They proposed a black box spoofing attack capable of deceiving all target detection models, achieving an 80\% success rate on the KITTI dataset. Both attacks~\cite{Cao2019a,Sun2020a}, necessitate precise pulse injection to generate adversarial examples and real-time knowledge of the victim LiDAR's position for maintaining a robust adversarial pattern while the vehicle is in motion. 

Hallyburton et al.~\cite{Hallyburton2022} investigated adversarial spoofing attacks on camera-LiDAR fusion in AVs. They introduced the 'frustum attack', a black-box technique exploiting ambiguities in multi-sensor fusion. The frustum attack enables the manipulation of LiDAR points while maintaining consistency with image detections. This is possible because a 2D camera image cannot fully resolve an object's 3D position. Moreover, the frustum attack demonstrates the capacity for consistent execution over time, forming stealthy longitudinal attack sequences that compromise the tracking module. However, the proposed attack requires that the attacker has prior knowledge of scene-specific information. Li et al.~\cite{Li2021} exploit the fact that LiDAR point clouds collected from a moving vehicle need calibration based on the moving trajectories, so they propose to spoof the vehicle’s trajectory with adversarial perturbations, which can distort the LiDAR sweeps and fool the 3D object detectors.  The paper describes a technique to manipulate the trajectory of 3D LiDAR data in nuScenes and deceive LiDAR-based perception modules. Motion compensation is a natural part of self-driving, and it has the potential to create an easily exploitable backdoor in deep networks used for perception and planning. To create adversarial spoofing, the method employs projected gradient descent (PGD) to generate small perturbations in the self-driving car's trajectory data.

In real-world autonomous driving scenarios, AVs are equipped with multiple sensors of multiple modalities e.g., LiDAR for point cloud data, a camera for RGB images, and more. This sensor redundancy enhances the reliability of estimations. Integrating multiple sensor data yields better environment perception, e.g., tracking dynamic vehicles. Advanced ML techniques employ multi-modal fusion to translate raw sensor data into meaningful semantic information. For example, Google Driverless Car fuses LiDAR with stereovision and Enhanced Maps for road scenery understanding. However, the emergence of multi-modal attacks in recent times has introduced an additional layer of complexity to the security challenges confronted by AVs~\cite{Tu2021,Cao2021,Abdelfattah2021,Wang2021}. Adversarial attacks on LiDAR-based AVs can involve cyber-attacks to inject manipulated point clouds or physical manipulation of the environment to deceive the system. Executing adversarial attacks in real-time usually requires precise pulse injection and continual tracking of the victim LiDAR's position to maintain robust adversarial patterns.

Attacks have been developed to perturb both camera and LiDAR inputs either separately~\cite{Tu2021} or concurrently~\cite{Cao2021}. Tu et al.~\cite{Tu2021} simulated adversarial attacks on multi-modal perception models by introducing an adversarial textured mesh that can be placed on a vehicle and make this vehicle invisible to the multi-modal perception models. Specifically, the adversarial mesh is first rendered into both LiDAR points and image pixels in a differentiable manner, and then the multi-modal inputs are passed through a fusion-based detector. Finally, an adversarial loss is employed to adjust the mesh parameters. Abdelfattah et al.~\cite{Abdelfattah2021} proposed a multi-modal adversarial attack on car detection models that rely on both LiDAR and RGB camera inputs. This attack employs a 3D adversarial mesh with a misleading texture, which is trained on the KITTI dataset. The deceptive texture tricks the 2D detection process, while the added LiDAR points in the point cloud create confusion for the 3D detection pipeline. The target models for their study are Frustum PointNet~\cite{Qi2018} and EPNet~\cite{Huang2020a}, which represent different approaches in multi-modal perception.

Additionally, researchers have utilized DNNs to integrate LiDAR point clouds with RGB camera images for monocular depth estimation (MDE), improving accuracy. MDE is a crucial aspect of scene understanding for AVs, aiding in object detection, collision avoidance, SLAM, and scene reconstruction. Point clouds can provide better depth information than cameras, but they are noisy and sparse. RGB images have limitations on maximum depth. Recently, researchers~\cite{Qiu2019,You2020,Wang2019} have utilized DNNs to integrate LiDAR point clouds with RGB camera images, improving the MDE accuracy. However, the underlying machine-learning models are vulnerable to optical signal attacks, especially localized attacks such as adversarial patches~\cite{Brown2017,Zhang2020a}. The adversarial patch deceives a DNN classifier into predicting any object, associating the pasted patch with a targeted class. This occurs irrespective of the object’s scale, position, or direction. These attacks have demonstrated effectiveness in a black-box scenario and can be printed and added to any scene. Cheng et al.~\cite{Cheng2022} investigated stealthy attacks on MDE using an adversarial patch optimization framework. The patch will automatically locate the highly effective area for attack on the target object and the impact of compromised MDE was observed on the downstream object detection task which could detect a benign object but not an adversarial one.

Current adversarial attacks targeting classification, object detection, and segmentation primarily focus on manipulating the probabilities assigned to various categories within the model, resulting in altered final predictions by the neural network. Nonetheless, these attacks exhibit limited impact on scene understanding models such as depth or distance estimation. Mathew et al.~\cite{Mathew2020} introduce a deep feature annihilation (DFA) loss for both perturbation and patch attacks. DFA loss deliberately corrupts the internal representation of a DNN, extending beyond the manipulation of final predictions.

\subsubsection{Poisoning Attacks}\label{322Poisoning}
Pretrained ML systems~\cite{Qi2017a,Qi2017b,Qi2018,Zhou2018,Autoware,Apollo,Hugging} are increasingly integrated into the AV pipelines (refer to Figure~\ref{fig_2}), primarily because of their reliance on extensive datasets that are computationally expensive to train and evaluate independently. While reusing pretrained models can alleviate resource constraints and reduce carbon footprints, it also introduces security risks~\cite{Jiang2022,Garcia2020,Gu2017}, particularly vulnerability to adversarial poisoning attacks~\cite{Wang2022a,Tian2022}. These attacks occur during the training phase, aiming to introduce biases and compromise the target model's overall performance, potentially inducing specific behaviors during inference. Adversaries may inject malicious samples into the training data~\cite{Goldblum2022,Carlini2023} or manipulate the model’s parameters~\cite{Rakin2021,Kurita2020}. Additionally, many developers and companies opt for Machine-Learning-as-a-Service (MLaaS) providers~\cite{Mund2015,Ravulavaru2018} to address the challenge of expensive training. While this simplifies the rollout of DNN-based LiDAR object detection models, it also exposes them to security vulnerabilities. Malicious actors involved may sabotage model performance by supplying tainted data or tampering with the training dataset, such as certain point cloud data providers, including those releasing public datasets~\cite{Geiger2013,Caesar2020,Huang2018}. Bishoff et al.\cite{Bishoff2023} researched on quantification of adversarial robustness of multispectral segmentation models against data poisoningin a universal black-box backdoor sample detection method tailored for 3D point clouds without any prior knowledge or assumption of the triggers and victim models.

Backdoor attacks~\cite{Li2024}, a subset of poisoning attacks, have drawn much attention in recent years. They involve injecting misleading triggers into the training dataset, prompting the model to learn specific patterns associated with them. Consequently, during inference, the model may make incorrect predictions when exposed to inputs containing these malicious triggers, while behaving normally on clean data samples. Due to their stealthiness,  backdoor attacks are challenging to detect and can impact all systems utilizing poisoned pretrained models, including those that may inadvertently contain intentionally poisoned samples from publicly available datasets.

Several works have explored backdoor attacks in 3D point clouds. Xiang et al~\cite{Xiang2021} proposed injecting triggers with optimal shape and size into training samples to fool point cloud classification models. Li et al.~\cite{Li2021a} exposed the vulnerability of 3D point cloud models to spatial transformations, employing a clean-label backdoor trigger attack using rotation-based feature disentanglement. This allows misclassification of objects with specific rotations, like a slightly tilted car being identified as a plant. Bian et al.~\cite{Bian2021} introduced a method for conducting stealthy 3D backdoor attacks by contaminating training data labels, ensuring trigger resistance to preprocessing operations like down sampling and anomaly detection. The authors claimed that even when the target position and size changes, or rotates, their label pollution remains highly effective. Considering the limitation of accessing original training data, Wen et al.~\cite{Wen2021} designed a backdoor attack method against 3D point cloud classification tasks using reverse engineering to generate 3D triggers and point cloud data.  

However, these studies primarily focus on object classification, neglecting the detailed information provided by object detection tasks, such as object locations, sizes, and orientations~\cite{Qian2022}. Moreover, these studies primarily explore backdoor attacks in simulated environments, raising concerns about their real-world effectiveness. Implementing physically realizable backdoor attacks against LiDAR object detection in AVs presents challenges due to disparities in model structures and the intricate nature of LiDAR point cloud data. Ensuring robustness against location errors and varying driving conditions is important for the effectiveness of such attacks. Also, maintaining the stealthiness of poisoned data during the training phase is essential to avoid detection in real-world settings.

Zhang et al.~\cite{Zhang2022} proposed a backdoor attack for LiDAR object detection systems using fake vehicle point clusters as hidden triggers during training. These triggers are difficult to detect due to sparse LiDAR signals. However, due to this sparsity, larger triggers, such as cargo carrier bags or exercise balls, are needed for effective capture. Mounting these large triggers on the target vehicle significantly alters its appearance, making 3D backdoor attacks unrealistic in practice. Chaturvedi et al.~\cite{Chaturvedi2023} introduced BadFusion, a 2D-oriented backdoor attack targeting LiDAR based multi-modal fusion systems for 3D object detection. BadFusion inserts fusion-aware 2D triggers into the camera signals to influence the LiDAR projections and manipulate the final predictions. They evaluated the attack against state-of-the-art fusion methods, achieving goals of resizing bounding boxes and causing objects to disappear.

Data poisoning attacks pose a significant threat to LiDAR systems and can cause more damage than the evasion attacks due to their attack success rates and stealthiness~\cite{Zhang2022}. Real-world 3D point cloud data is often noisy, with occlusions and spatial deformations, making it easy for attackers to hide malicious triggers. Furthermore, the coordinate-based representation and irregular sampling of point clouds make it difficult to verify data integrity and detect malicious triggers, complicating mitigation efforts.

\subsubsection{Model Stealing and Privacy Attacks}\label{323Stealing}
Model stealing, also known as model extraction, aims to obtain various components of a black box ML model's architecture, such as training hyper-parameters or learned parameters. In this attack, adversaries replicate the target model's functionality by querying it with crafted inputs and using the responses to train a surrogate model~\cite{Jagielski2020,Wang2018,Genc2023}.  In LiDAR-based perception systems, an attacker could attempt to steal the underlying models used for tasks like object detection, segmentation, or localization. However, the feasibility and effectiveness of these depend on various factors, including the complexity of the models involved, the availability of data for training the surrogate model, and the attacker's knowledge of the target system. As previously discussed in Section~\ref{322Poisoning}, MLaaS is widely adopted in autonomous perception systems, enabling users to bypass time-consuming tasks like data collection, hyper-parameter tuning, and model training. Oliynyk et al.~\cite{Oliynyk2023} addressed model stealing threats in MLaaS, categorizing stealing attacks, evaluating defenses, and providing guidelines for strategy selection.

Apart from stealing the model, attackers also engage in inference attacks to access underlying data. These attacks come in two main forms: model inversion attacks~\cite{Fredrikson2015,Zhang2020} which reconstruct or approximate the model’s original training or input data by leveraging its outputs or architecture, and membership inference attacks~\cite{Shokri2017,Hu2022}, which analyze the model’s outputs to determine if a specific data point was part of its training dataset. Pittalunga et al.~\cite{Pittaluga2019} presented a privacy attack that can reconstruct the scene by inverting sparse point cloud models. Pan et al.~\cite{Pan2019} studied membership inference attacks on Deep Reinforcement Learning (DRL) within navigation tasks using LiDAR perception as input. Their attack aimed to extract information about the dynamics of the model training environment, as DRL policies may inadvertently disclose sensitive information. DRL algorithms are being deployed in LiDAR based navigation systems to make complex decisions in dynamic environments by training agents through trial and error~\cite{Huang2021}.  These agents continuously make decisions by assigning rewards to actions and optimizing policies for action selection. 

As transferable attacks become more prevalent and complex, there's an increased likelihood of model stealing and privacy attacks becoming more common in LiDAR-based systems. While these systems may exhibit resilience against certain attacks, security remains an ongoing concern in autonomous vehicles (AVs). Attackers have the potential to develop new strategies or exploit vulnerabilities in different system components. Hence, security measures and research into robustness are continually evolving to mitigate potential threats.

\section{Defenses}\label{4Defenses}
The field of autonomous driving using 3D LiDAR requires robust sensor-based perception systems to ensure safe and reliable operation. Researchers have explored various defense strategies to protect against adversarial attacks, including cyber-attacks targeting autonomous ML systems, but there has been limited work related to 3D LiDAR-based autonomous driving systems. In this section, we review studies that address several aspects of sensor-based perception defense. While these studies provide valuable insights, some limitations remain in terms of comprehensive validation and applicability within the context of autonomous driving. We explore each study's findings, highlighting their contributions and potential avenues for further exploration.

\subsection{Defenses for Sensor Attacks}\label{41Defenses}
Whether utilizing 3D data or not, autonomous driving technologies expose vehicles to potential cyber-attacks, prompting the adoption of countermeasures by AVs to mitigate cyber threats. For example, the attacker can inject false information or biased data into the network, leading to incorrect conclusions or decisions in its ML systems. The Sybil attack is among the attacks that involve broadcasting false information to AV network nodes, potentially disrupting the proper functioning of ML-based driving systems. Authentication and verification of sensor data sources, such as implementing digital signatures, timestamps, encryption, synchronization, etc., contribute to the defense against Sybil attacks~\cite{Zhang2008,Chen2009,Hao2011,Hasan2020,Lim2020}. Also, graph-based Sybil detectors~\cite{Shoukry2018,Wang2016} are employed to identify and detect Sybil attacks within network or graph-based systems. These detectors operate based on the analysis of physical or logical proximity within the graph. In the context of Sybil attacks, malicious nodes may attempt to manipulate the structure of the proximity graph, which can result in inconsistencies detectable by the defense mechanism.

Amoozadeh et al.~\cite{Amoozadeh2015} investigated vulnerabilities within localization and navigation technologies that adversaries could potentially use to manipulate AV navigation. The paper introduced multiple countermeasures, including detection, filtering, encryption, digital signature, and others, to mitigate these weaknesses, addressing threats like spoofing attacks, replay attacks, Sybil attacks, and message falsification attacks.

Porter et al.~\cite{Porter2021} introduced a proof of concept for detecting replay attacks on sensors in autonomous vehicles. They employed a technique called linear time-varying dynamic watermarking to achieve this. This method involves injecting a unique excitation into control inputs to safeguard measurement signals, and it has been demonstrated to effectively detect generalized replay attacks, both theoretically and through simulations in the context of AVs.

Multiple studies have examined ML-based approaches including Intrusion Detection Systems (IDS) utilizing anomaly detection techniques to detect Sybil attacks~\cite{Chen2009,Luo2021,Helmi2022,Manale2022}. Helmi et al.~\cite{Helmi2022} involved analyzing Sybil attack patterns, considering factors such as time, position, and traffic density. Mechanisms that detect disparities in motion trajectories and, hence, identify potential Sybil attacks, can also be applied in the context of AVs~\cite{Chen2009}. Luo et al.~\cite{Luo2021} introduced a detection mechanism based on Temporal Graph Convolutional Networks, leveraging the fact that inconsistencies in perception and communication can be used to detect attacks. 

These defensive mechanisms, including anomaly detection approaches, do not explicitly address AML attacks in autonomous driving systems. However, it is important to note that cyber defensive approaches designed to counter cyber-attacks could also provide a solution to address AML attacks, regardless of whether those defensive approaches were initially developed to combat AML attacks.

When addressing defenses against 3D-based attacks such as spoofing attacks, some researchers~\cite{Petit2015,Yan2016} advocate introducing randomness into data collection to introduce complexity for potential malicious attackers, assuming a fixed probe window for LiDAR signal reception. An alternative approach suggested by others~\cite{Petit2015,Yan2016,Lim2018,Hallyburton2022} involves leveraging multiple sensors, including LiDAR, with data from these sensors being fused into the input.  Incorporating diverse sensors, such as cameras and LiDARs, helps address challenges like object omissions and extended image processing due to distortions.

While these methods can improve the overall system performance and reliability, they do not specifically address the robustness of ML systems within the autonomous driving context. Robustness in ML systems typically involves ensuring that the system can handle unexpected or adversarial inputs without compromising its functionality, which goes beyond the methods discussed in the referenced papers. 

There is some research on the robustness of autonomous driving systems using LiDARs. Hau et al.~\cite{Hau2021} introduces Shadow-Catcher, a defensive method designed to identify LiDAR spoofing attacks that can deceive 3D object detection systems. The fundamental insight behind this approach is the distinction between the 3D shadows of genuine objects and those of spoofed objects. Shadow-Catcher employs ray optics to map the anticipated shadow region of a detected 3D object. If an abnormal shadow is detected, it employs a binary classifier that relies on density features extracted from the proposed shadow region to categorize it as either a ghost object shadow or a poisoned shadow. This classification helps verify the existence of real objects and allows the system to identify fake objects, but solely when their shadows exhibit noticeable distinctions from those of genuine objects. Considering that the effectiveness of systems which use the information of shadow could be impacted by adverse weather conditions or poor lighting, the accuracy of shadow prediction in Shadow-Catcher may be affected by external factors. This approach might not be universally applicable to all types of LiDAR spoofing attacks as it focuses exclusively on the detection of fake object insertions.

Hau et al.~\cite{Hau2022} also utilize shadow information, presenting a method different from Shadow-Catcher. Their approach aims to detect objects concealed by an adversary within void regions in 3D point cloud scenes. It determines whether these void regions are caused by occluded detected objects or by unidentified obstacles, potentially indicating adversarial hidden objects. Even when an object is intentionally concealed from traditional 3D object detectors, this method can identify it by searching for void spaces in the front-near region of the ego-vehicle, which are indicative of shadow regions. Using ray frustums originating from the LiDAR to the shadow clusters, they generate frustums and identify points falling within them. In this method, there is a possibility of false positives, where void regions are mistakenly attributed to adversarial hidden objects when they are due to other factors, such as LiDAR noise or sensor limitations. Also, in scenes with a high density of objects, shadows from multiple objects may overlap, making it challenging to accurately distinguish shadows from individual obstacles. Further, an object may not be detected if it is hidden in a way that it does not cast a shadow.

You et al.~\cite{You2021} propose an approach that leverages motion as a physical invariant of genuine objects to detect spoofing attacks. The approach called 3D Temporal Consistency Check (3D-TC2) utilizes spatial-temporal information derived from motion prediction to validate objects identified by 3D object detectors. 3D-TC2 relies on motion as a key invariant. Thus, if an object is stationary or has limited motion, the approach might struggle to detect spoofing attacks effectively in such scenarios. In addition, training the model to recognize genuine motion patterns could require a substantial amount of diverse and representative data, which might be difficult to obtain in all scenarios. 

Sato et al.~\cite{Sato2023} explore various aspects of the next-generation LiDARs’~\cite{Yoshioka2022} security measures, including laser-timing randomization, pulse fingerprinting, and simultaneous emissions. Previous studies have predominantly used the earlier version of LiDARs such as VLP-16~\cite{VelodyneLiDAR}, emphasizing synchronized attack scenarios. The authors highlight the challenge of implementing spoofing attacks on the next-generation LiDAR systems with simultaneous laser firing, as returning laser pulses to each simultaneous laser becomes impractical. It also describes other security aspects of the next-gen LiDAR system and its limitations. For instance, timing randomization in next-generation LiDARs enhances security by making it challenging for attackers to synchronize with laser firing patterns. However, some limited point injection remains possible if the randomization is not sufficiently strong. Existing spoofing attacks may not yield the anticipated effectiveness; however, existing attacks can still achieve a success rate exceeding 35\% on industry-grade detectors even when employing timing randomization using the next-generation LiDARs. Pulse fingerprinting, while effective, is not entirely foolproof against spoofing attacks, as certain laser-firing frequencies can still coincide with the fingerprinting interval, potentially enabling limited spoofing. Additionally, the use of laser wavelengths outside the LiDAR's detection range, as observed in rare cases, may serve as a mitigation strategy against spoofing attacks. This strategy involves employing laser wavelengths that fall beyond the LiDAR's typical operating range, making it challenging for attackers to manipulate the LiDAR system effectively using these unconventional wavelengths. However, it is important to note that not all next-generation LiDAR systems necessarily employ this strategy, and this approach may have limited effectiveness. 

\subsection{Defenses for ML Attacks}\label{42Defenses}
In the context of defending autonomous systems against adversarial inputs such as adversarial attacks and adversarial spoofing attacks, several strategies have been proposed, including input transformations, adversarial training, and certified robustness. However, most of these techniques have primarily focused on image classification models when dealing with digital-space attacks, rather than addressing the challenges posed by object detection models operating in real-world settings.

Giannaros et al.~\cite{Giannaros2023} explore threats such as relay signal attacks and spoofing attacks in AVs using LiDAR. They also briefly mention theoretical adversarial attacks which can craft objects that appear ordinary to human observers but are wrongly classified by the LiDAR system. The authors explain defensive methods such as blockchain in AVs against cyber-attacks and recognize that it is challenging to find truly effective defense mechanisms against adversarial attacks. 

Sun et al.~\cite{Sun2020} focused on stereo-based 3D object detection, where 3D spatial information is extracted from left and right images. They introduced a defensive approach using adversarial training, incorporating a regularization term defined as the disparity between the differences in the original left/right images and the differences in the attacked images. While it demonstrates reliable performance, the analysis is limited to the robustness against commonly used adversarial attacks including FGSM~\cite{Goodfellow2014} and PGD~\cite{Madry2018} attacks which are designed to manipulate image classifiers, not for autonomous driving systems. Also, adversarial training might not be suitable in this domain since it primarily focuses on classification models and requires the use of norm-bounded perturbations.

Cao et al.~\cite{Cao2019a} outlines a multi-tiered defensive strategy against spoofing attacks, which aim to deceive machine learning models during object detection in AVs. In the context of autonomous systems, the attacks primarily exploit ground-reflected points within falsified obstacle point clouds and their subsequent transformation into a 2D matrix, resulting in data loss. To counter these threats, the authors suggest several defensive measures such as excluding ground reflection data points, avoiding the 3D-to-2D conversion, and increasing the number of features. At the sensor level, they propose sensor fusion, modifying LiDAR structures, and employing randomization methods, albeit at the potential cost of system performance. Additionally, at the ML level, they recommend adversarial training. However, it is important to note that the paper lacks comprehensive analysis and empirical or theoretical validation to confirm the effectiveness of these defenses in the specific context of autonomous driving.

Various defense methods exist for countering adversarial attacks in point cloud classification, including adversarial training~\cite{Sun2021,Wang2022}, Gaussian noise perturbation~\cite{Yang2019}, certified robustness~\cite{Liu2021} and point removal~\cite{Liu2019,Zhou2019}. Zhou et al.~\cite{Zhou2019} proposed DUP-Net, leveraging statistical outlier removal (SOR) and up-sampling to enhance models’ resilience to adversarial corruptions. Although SOR effectively eliminates adversarial points, it lacks full robustness when confronted with incomplete partial point clouds. Concurrent research~\cite{Ren2022,Yu2023,Sun2022} evaluates the robustness of point cloud classifiers against adversarial corruptions.

Deng et al.~\cite{Deng2021}  discusses defenses against adversarial evasion attacks and poisoning attacks in sensor-based AV systems, classifying them into proactive and reactive methods. Proactive methods enhance deep learning model robustness, including adversarial training, network distillation, regularization, ensemble, and certified defense. Reactive approaches focus on detecting and countering adversarial examples before model input. The study noted that the existing research primarily focused on evaluations within the realm of image classification, and current adversarial defensive methods may not be suitable for AVs. For example, proactive methods like adversarial training demand large datasets and substantial training time. In addition, model ensemble methods require the expense of additional resources. On the other hand, the effectiveness of integrating network regularization and robustness methods into the training process in complex autonomous driving systems needs further verification. Reactive methods may degrade performance on normal inputs, which is unacceptable for safety-critical AVs. Moreover, adversarial detection techniques should be further explored. Real-time monitoring and defense are crucial for maintaining the safety of AVs, given the real-time nature of autonomous driving processes. While the study does touch upon attacks in autonomous driving systems and the potential adaptation of defensive concepts to this field, it falls short of providing a thorough analysis or substantial experimental/theoretical validation for such applications. Furthermore, it lacks an analysis of machine learning methodologies in autonomous driving systems that are vulnerable to these attacks and could potentially benefit from these defenses. 

Sun et al.~\cite{Sun2020a} introduced two autonomous driving architectures, oCclusion Aware hieRarchy anomaLy detectiOn (CARLO) and Sequential View Fusion (SVF). These architectures defend the systems against LiDAR spoofing attacks, especially constrained by the sensor attack capability. The authors observed that the breach of the occlusion's physical law is a common factor facilitating LiDAR spoofing attacks. Consequently, CARLO utilizes occlusion patterns as inherent physical constants to detect spoofed fake vehicles. SVF processes raw input point clouds by removing outliers and converting them into a front-view (FV) representation using a point cloud segmenter. It calculates segmentation scores for each point, merges them with input features, and feeds the augmented data into the 3D object detector. The framework assesses resilience against adversarial attacks in semantic segmentation and 3D object detection. However, the attacks applied in the paper are limited and the defensive approaches they proposed cannot be universally applied to autonomous driving systems.

Much like the defense against cyber-attacks, one defense strategy against physical adversarial attacks involves incorporating additional perception sources~\cite{Cao2021,Zhu2021}, such as multiple cameras and LiDARs, even if they share overlapping views from different positions. However, sensor fusion increases the cost and complexity of AV systems.

In a study conducted by Cao et al.~\cite{Cao2021}, they examined the impact of adjusting input data quality such as bit-depth reduction, median smoothing, JPEG compression and auto-encoder reformation, and utilizing adversarial training as defense measures. While these defense strategies have shown their effectiveness in image space, they reduced the success rate of attacks to only to 66\%. This outcome indicates the need for more robust defense mechanisms in this domain. Additionally, it explores the inclusion of radar as a perception source for defense, as it may enhance security if the radar perception model proves to be robust. The authors acknowledge that potential defense strategies necessitate further investigation in their forthcoming research.

Xiang et al.~\cite{Xiang2019} observed that 3D adversarial examples tend to have limited transferability across different models, meaning that if one model falls victim to an attack, others can still produce accurate results. Based on this insight, Zhu et al.~\cite{Zhu2021} explore the concept of output aggregation. This strategy involves combining outcomes from multiple point cloud segmentation models, each trained with distinct data augmentation and variable initialization. However, the success of output aggregation critically hinges on the diversity of the models employed, which can introduce a trade-off against benign accuracy.

Tu et al.~\cite{Tu2020} employed meshed objects, such as canoes and couches, placed on car rooftops to improve the defense of LiDAR object detection models.  By subjecting these meshes to an adversarial attack and augmenting them into the AV training data, they improved the detector’s robustness and its ability to generalize to unseen samples. However, their method is limited to specific roof-mounted objects, lacking broader applicability.  In contrast, Lehner et al.~\cite{Lehner2022} proposed the 3D-VField method to achieve stronger generalization by utilizing perturbed objects for data augmentation without assumptions on the object type. They improved realism by considering occlusion constraints typically overlooked by other defense strategies, making perturbations sensor-aware and applicable across various indoor and outdoor environments. To defend against privacy attacks through membership inference or model inversion, Liu et al.~\cite{Liu2023} proposed a privacy protection method for 3D point clouds based on an optical chaotic encryption scheme making it unclassifiable and unidentifiable, while decryption retains the original data.

Even though these studies have shown several defensive approaches against multiple types of adversarial attacks, none of them has covered the range of possible adversarial attacks described in Section~\ref{3Attacks}, which can occur within the realm of autonomous driving that involves sensor data. That is, their defensive approaches have limited applicability to such applications. Therefore, there is still a significant research gap on the topic of defensive approaches in autonomous driving.

\section{Discussion}\label{5Discussion}
Securing the integrity of autonomous vehicle’s ML components is paramount for ensuring its safe and efficient operation. This paper places a primary focus on adversarial attacks directed at the vehicle's LiDAR system and its underlying ML algorithms. Such attacks are known for their stealth and deception, posing a substantial threat. Their consequences range from temporary disruptions in the AV data flow to, in extreme cases, the potential for sudden collisions. Our survey of adversarial attacks and defenses on LiDAR for AVs highlights the necessity for understanding and countering these threats. The attacks on perception module have a cascading effect on the decision making component. For example, if an adversarial attack distorts the perception of nearby obstacles, the decision-making component may fail to appropriately navigate around them, potentially resulting in accidents or collisions.

Researchers have explored AML attacks manipulating point cloud location and quantity. However, translating these attacks into practical scenarios poses substantial challenges. Real-time execution is hindered by the dynamic movement of the victim AV and the precision required to target the LiDAR sensor effectively. Furthermore, specialized equipment is essential for generating laser signals, reducing the flexibility of such attacks. Additionally, adversarial objects used in some attacks require specific shapes, limiting their flexibility and potentially raising suspicion due to their abnormal appearance to human observers. Moreover, LiDAR scanning errors, manufacturing errors, and location inaccuracies can distort point cloud shapes, rendering the attack ineffective. Furthermore, LiDAR sensors with fixed angular frequencies impose constraints on spatial resolution and point capture capacity per scan, potentially restricting the feasibility of achieving arbitrary point cloud perturbations. These findings underscore the importance of understanding the physical limitations of LiDAR sensors in the context of adversarial attacks. 

Most of the cited attacks assume that the attacker possesses full knowledge of the model's architecture and parameters. Although these scenarios are instrumental in pushing the boundaries of current techniques and comprehending their weaknesses, they do not accurately reflect the security challenges encountered in real-world settings, where attackers may lack access to such detailed parameters. While attacks may target specific models or systems, their effects can spread through transferable attacks, which remain poorly understood, particularly in cyber-physical security contexts. Recently, Wang et al.~\cite{Wang2023} categorized and summarized transferable attack designs, focusing on their impact on systems like autonomous driving, speech recognition, and LLM systems, offering insights for comparing attack types across domains. Moreover, a transferable attack on a multimodal model involves crafting attacks effective for multimodal data and models, exploiting vulnerabilities in cross-modal fusion mechanisms, and emphasizing the need for robustness against various data types in multimodal systems.

Also, there are no robust and effective solutions against all possible adversarial threats, so we need to understand the limitations of existing defenses, both in theory and in practice. For example, threats are constantly evolving, and adversaries are becoming more sophisticated. Defensive techniques that work against known attacks may struggle to adapt to unknown novel attacking tactics. This dynamic situation requires continuous updates to defensive mechanisms. Although autonomous vehicles heavily rely on their perception systems, comprising various sensors and complex ML models, compromising a single element may not immediately lead to incorrect decisions. However, when integrated with data from other sensors, adversarial interventions can significantly compromise safety~\cite{Wang2021}. This underscores the importance of well-designed defense strategies.

Autonomous vehicles generate and process massive amounts of sensor data in real-time. Implementing effective defenses at scale, especially for complex data sources like LiDAR, can be computationally intensive and resource demanding. Balancing scalability without compromising system performance is a major challenge. Furthermore, AVs operate in diverse environments and weather conditions. Defensive mechanisms need to adapt to these environmental variabilities, ensuring that they do not generate false alarms.

Furthermore, as AVs gain more autonomy, ethical considerations about defense strategies may arise. For instance, envision a scenario where an autonomous vehicle must respond to an AML attack that poses a potential threat to pedestrians. Even in the context of defending against such attacks, the vehicle’s decisions must align with established ethical norms and societal expectations, with a primary focus on upholding human safety standards. Prioritizing human safety not only underscores the ethical responsibility associated with autonomous vehicles technology but also serves as a crucial guiding principle in the development and implementation of robust defense mechanisms against adversarial attacks in LiDAR-based systems. Further exploration into prioritizing human safety within these frameworks constitutes an essential avenue for future research. 



\begin{IEEEbiographynophoto}{Junae Kim}
	Junae Kim is a research specialist at Defence Science and Technology in Australia. Her research interests include adversarial machine learning, ML-based autonomous systems and the robustness of ML. Kim received her PhD in machine learning from the Australian National University, Australia. She is an honorary scholar at the University of Queensland. Contact her at junae.kim@defence.gov.au. 
\end{IEEEbiographynophoto}

\begin{IEEEbiographynophoto}{Amardeep Kaur}
	Amardeep Kaur is a senior researcher at Defence Science and Technology in Australia. Her research interests include generative models, AI trustworthiness, and anomaly detection algorithms. Kaur received her PhD in data mining from the University of Western Australia. Contact her at amar.kaur@defence.gov.au.
\end{IEEEbiographynophoto}

\vfill

\end{document}